\RequirePackage{fix-cm}
\documentclass[final]{svjour3}       
\usepackage[compact]{titlesec}
\smartqed  
\usepackage[margin=3cm]{geometry}
\usepackage[linesnumbered]{algorithm2e}
\usepackage{graphicx}
\graphicspath{ {./Figures/} }
\usepackage{caption, copyrightbox}
\usepackage{subcaption}
\usepackage{booktabs} 
\usepackage{xparse}   
\def\rot{\rotatebox}

\usepackage{enumitem}

\usepackage{geometry}
\usepackage[fontsize=12pt]{fontsize}

\usepackage{algpseudocode}
\def\makeheadbox{\relax}
\usepackage{gensymb}
\usepackage{amsmath}
\usepackage{amssymb}
\usepackage{hyperref}
\newcommand\tab[1][1cm]{\hspace*{#1}}
\newcommand{\D}{\mathbb{D}}
\newcommand{\V}{\mathbb{V}}
\newcommand{\F}{\mathbb{F}}

\newcommand{\E}{\mathbb{E}}
\renewcommand{\S}{\mathbb{S}}
\newcommand{\Z}{\mathbb{Z}}
\newcommand{\collection}[1]{\ensuremath{\left\{#1 \right\}}}
\newcommand{\fof}[1]{\ensuremath{\left(#1\right)}}
\DeclareMathOperator*{\argmax}{argmax}

\begin{document}


\title{A Novel Factor Graph-Based Optimization Technique for Stereo Correspondence Estimation
}
\author{}
\subtitle{Hanieh Shabanian, Madhusudhanan Balasubramanian}

\titlerunning{Factor Graph-based Stereo Disparity}        



\institute{H. Shabanian \email{hshbnian@memphis.edu} \and M. Balasubramanian \email{mblsbrmn@memphis.edu} \\ Department of Electrical and Computer Engineering, The University of Memphis} 

\date{}

\maketitle
\vspace{-0.5in}
\begin{abstract} 

Dense disparities among multiple views is essential for estimating the 3D architecture of a scene based on the geometrical relationship among the scene and the views or cameras.  Scenes with larger extents of heterogeneous textures, differing scene illumination among the multiple views and with occluding objects affect the accuracy of the estimated disparities.  Markov random fields (MRF) based methods for disparity estimation address these limitations using spatial dependencies among the observations and among the disparity estimates.  These methods, however, are limited by spatially fixed and smaller neighborhood systems or cliques.  In this work, we present a new factor graph-based probabilistic graphical model for disparity estimation that allows a larger and a spatially variable neighborhood structure determined based on the local scene characteristics.  We evaluated our method using the \textit{Middlebury benchmark stereo datasets} and the \textit{Middlebury evaluation dataset version 3.0} and compared its performance with recent state-of-the-art disparity estimation algorithms.  The new factor graph-based method provided disparity estimates with higher accuracy when compared to the recent non-learning- and learning-based disparity estimation algorithms. In addition to disparity estimation, our factor graph formulation can be useful for obtaining \textit{maximum a posteriori} solution to optimization problems with complex and variable dependency structures as well as for other dense estimation problems such as optical flow estimation.


\end{abstract}

\section*{Declarations}
    \underline{Funding}: This research was supported in part by an unrestricted start-up fund from the Herff College of Engineering and a graduate assistantship from the Department of Electrical and Computer Engineering, The University of Memphis.\\
    \underline{Conflicts of interest}: None.\\
    \underline{Availability of data and material}: Public benchmark datasets\\
    \underline{Code availability}: Available upon request

\section{Introduction}
\label{intro}
The 3D architecture of an object or a scene can be estimated from two or more views of the scene by determining dense correspondences among the multiple views of the scene. A disparity map comprised of dense pixel-level correspondences between images in a stereo pair can be estimated using stereo matching algorithms. Stereo disparity estimation has wide-ranging applications such as robot navigation \cite{desouza2002vision}, aerial data analysis \cite{svensk2017evaluation}, image sequence analysis \cite{huang2008binocular}, and 3-D surface reconstruction \cite{remondino2008turning}. The presence of large untextured regions (homogeneous intensity), occluding objects and uneven intensity distributions in the stereo pair of images introduce significant challenges in estimating stereo disparity maps.

Two broad categories of stereo matching methods are window-based and energy-based algorithms \cite{scharstein2002taxonomy}. The energy-based algorithms \cite{gu2008local} are global methods with a cost function defined as a function of the entire image extent \cite{yang2008stereo}. Window-based or local methods utilize  a finite support window to define the cost function and are suitable for real-time applications. However, regions with homogeneous texture and occlusion affect the accuracy of the disparity estimated using local methods.

In general, stereo disparity estimation procedures include an initial cost calculation step, a cost aggregation step, an optimal disparity estimation step, and a disparity refinement step. Some of the successful and commonly used disparity cost measures are normalized cross-correlation (NCC) \cite{roma2002comparative}, sum of squared differences (SSD) \cite{marghany20113d}, sum of absolute differences (SAD) \cite{parvathysurvey}, gradient-based measures \cite{de2011stereo}, feature-based measures \cite{wang2014feature}, rank transform (RT) \cite{gac2009high}, and census transform (CT) \cite{mei2011building}.  Cost aggregation is useful for minimizing matching uncertainties and improving the accuracy of the estimates \cite{fang2012accelerating}.  During cost aggregation, initial disparity cost measures within the support region of each pixel are aggregated \cite{hamid2020stereo} such as using a low pass filter with a fixed kernel size or a variable support window (VSW) \cite{tombari2008classification}, using adaptive support weight (ASW) \cite{yoon2006adaptive}, and using a cross-based support window \cite{zhang2009cross}. Using edge-preserving filters such as a bilateral filter (BF) \cite{yoon2006adaptive,zhu2015edge} and guided image filter (GIF) \cite{zhu2016edge} for cost aggregation provides a significant improvement on the final disparity estimates over the initial disparity estimates. An optimal disparity map or disparity estimates are obtained from the aggregated cost volume using optimization procedures such as the winner-take-all (WTA) \cite{chang2018real} or using global optimization procedures such as dynamic programming(DP) \cite{arranz2012multiresolution}, graph cut (GC) \cite{wang2013effective}, intrinsic curves \cite{tomasi1998stereo}, and belief propagation \cite{liang2011hardware} methods. 

Establishing dense correspondences among geometrical coordinates of multiple views is an ill-posed problem due to scene occlusion.  In addition, larger scene extents with smoother texture characteristics and differences in scene illumination with respect to the multiple views/observers are sources of difficulties in disparity estimation.  A successful strategy for improving the accuracy of the disparity estimates is to utilize the spatial dependencies of the scene characteristics as well as of the disparity estimates.  Among the probabilistic inference formulations for estimating dense geometric correspondences \cite{sun2003stereo}, Markov Random Fields (MRF) based approaches have been successful in modeling spatial geometrical dependencies \cite{kim2015error}. In brief, unknown true disparities among multiple views of a scene are modeled as random variables on the pixel lattice (random field) and dependence among random variables is assumed to follow Markov property. Thus, MRF represents a joint distribution of the random field using conditional distributions of each of the random variables.  In learning-based MRF models, parameters of the MRF potential functions are either learned separately from training data \cite{lan2006efficient,roth2009fields} or along with the unknown states of the random variables \cite{zhang2005parameter}. 

One of the limitations of the MRF models is that the neighborhood system used for enforcing spatial dependencies needs to be \textit{maximal}.  Further, the chosen dependency structure is uniformly enforced for all the random variables.  Therefore, pairwise cliques or $2 \times 2$ cliques are more commonly used in MRF models.  While learning methods are available for optimizing MRF parameters and neighborhood structure, they are generally limited to specific tasks.

In this paper, we present a new factor graph-based probabilistic graphical model (FGS algorithm) for disparity estimation that addresses the aforementioned MRF limitations.  Specifically, our model allows a larger neighborhood system and a spatially variable neighborhood structure dependent on the local scene characteristics. The proposed factor graph framework can also be used for solving general-purpose optimization problems from its posterior distributions.  Further, we present strategies for reducing the computational cost and accelerating convergence of factor graph messages namely by using \textit{a priori} disparity distributions with smaller support and using factor node potential functions that significantly reduce marginalization calculations. \textit{A priori} disparity probabilities were estimated using a previously developed disparity cost calculation framework \cite{Shabanian2021Hybrid}. We demonstrate the performance of the proposed probabilistic factor graph model by conducting extensive experiments using the Middlebury benchmark stereo datasets \cite{scharstein2003high}, \cite{scharstein2007learning}, \cite{hirschmuller2007evaluation}, \cite{scharstein2014high} and compare its performance with other state-of-the-art disparity estimation algorithms using Middlebury evaluation dataset version 3.0 \cite{scharstein2002taxonomy}.

The remainder of the paper is structured as follows. In Section~\ref{sec:FG model}, we present a detailed description of the new probabilistic factor graph-based stereo disparity estimation (FGS) algorithm.  We present our experimental results in Section~\ref{Experiments} and conclude this work in Section~\ref{Conclusions}.
\section{Probabilistic Factor Graph Model for Disparity Estimation}
\label{sec:FG model}
For notational convenience, pixel coordinates of images of size $M \times N$ are referred using a linear index $i \in \collection{1, \ldots, MN}$.  Rectified stereo image pairs were corrected for any illumination differences using a homomorphic filter \cite{oppenheim2004frequency}.  In brief, each image $I(i)$ is modeled as an interaction between an illumination component $l(i)$ and a reflectance component $r(i)$ as $I(i) = l(i) \, r(i)$.  Assuming that the illumination varies gradually over the imaging area, the illumination variation is subtracted from each image in the $\log$ domain using a high-pass filter. For each of the rectified and illumination corrected stereo image pairs, disparity was calculated with respect to the left image channel (reference image).
\subsection{Graph Structure and Message Passing for Approximate Inference}
\label{sec:FG structure}
\begin{figure}
    \includegraphics[width=0.70\textwidth,keepaspectratio]{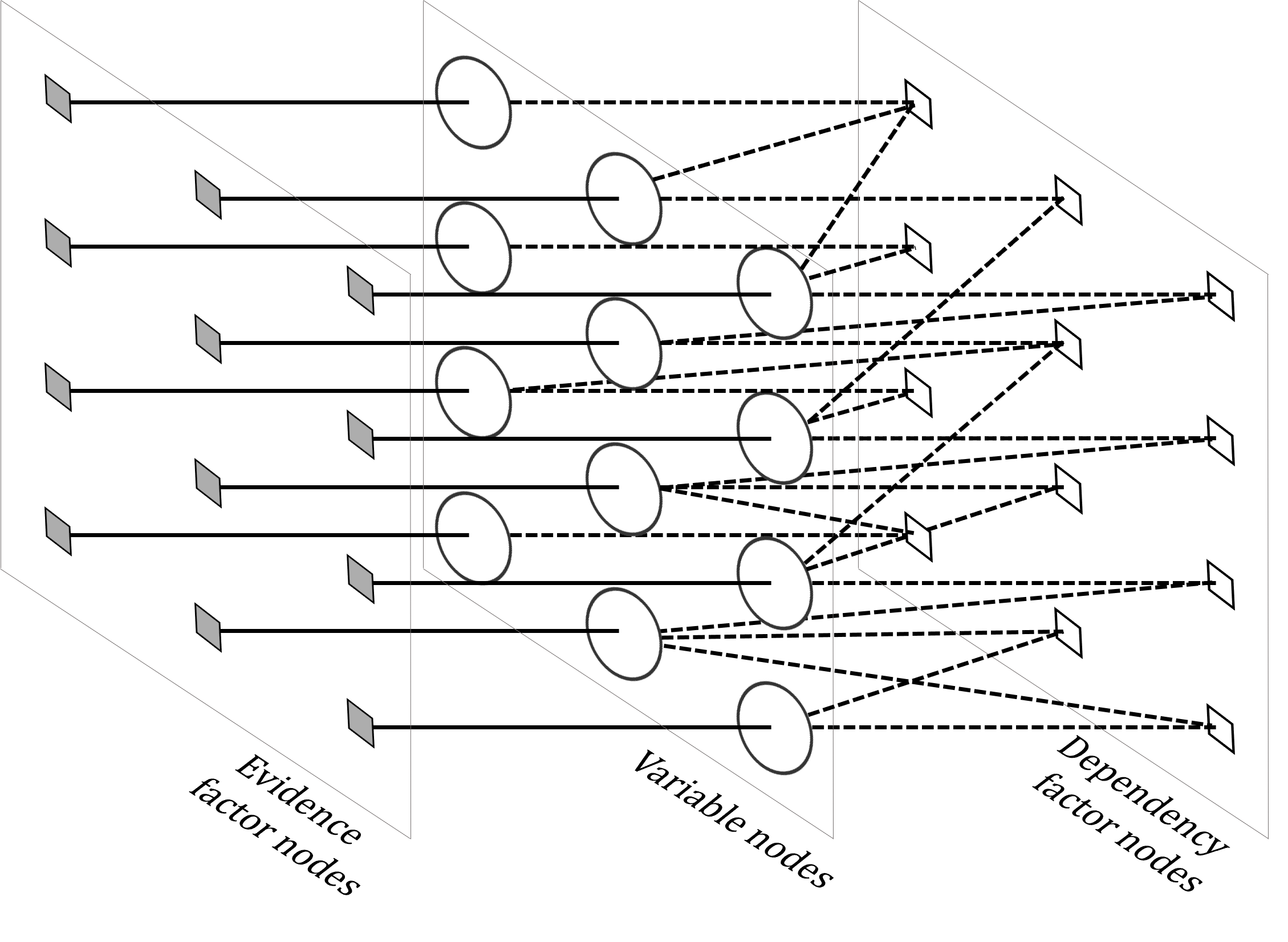}\centering
    \caption{Schematic diagram of the proposed factor graph model (FGS algorithm) designed for optimal estimation of dense stereo disparities. For images of size $M \times N$, there are $MN$ number of variable nodes (circular nodes) $\V$, $MN$ number of dependency factor nodes (empty square nodes) $\S$ and $MN$ number of evidence factor nodes (solid square nodes) $\E$.  The variable nodes represent posterior disparities to be estimated; the evidence factor nodes represent prior information about the dense disparities; and the dependency factor nodes represent the likelihood of satisfying spatial dependencies given localized disparity estimates.}
    \label{fig:FG Disparity}       
\end{figure}

Figure~\ref{fig:FG Disparity} shows the schematic diagram of the factor graph model designed for an optimal estimation of disparities $\D = \collection{d_i: d_i \in [d_{\min}, d_{\max} ] \subset \Z}_{i=1}^{MN}$ at each of the image pixel locations $i \in \collection{1, \ldots, MN}$.  The bipartite graph is comprised of a set of variable nodes $\V$ and a set of factor nodes $\F = \E \cup \S$.  The variable nodes $\V = \collection{1, \ldots, MN}$ represent disparity labels assigned to each pixel.  Evidence factor nodes $\E = \collection{1, \ldots, MN}$ provide prior degree of belief or evidence in assigning possible disparity labels at each pixel location.  Dependency factor nodes $\S = \collection{1, \ldots, MN}$ are used to model spatial dependencies among the disparity labels assigned to the neighboring pixels. 

A random variable $d_i \in \D$ is assigned to each \textit{variable node} $i \in \V$ to represent the disparity label assigned to the $i$th pixel location.  Each $j$th \textit{evidence factor node} in $\E$ is connected one-to-one with the corresponding $i$th variable node in $\V$ to incorporate prior belief or evidence in determining the disparity labels $d_i$.  To represent the influence of each pixel location on its neighboring pixels, each variable node $i \in \V$ is connected to one or more \textit{dependency factor nodes} $\collection{k \in \S}$ using localized intensity characteristics in the reference image as described in Section~\ref{sec:filters}.  

Let, $ne(i)$ represent the set of neighboring nodes connected with any given node $i$; $ne(i)\setminus j$ represent a set of all neighboring nodes of $i$ excluding node $j$; $\D_j = \collection{d_i, \forall i \in \V : i \in ne\fof{j \in \F}}$ be a collection of random variables associated with any factor node $j \in \F$; and let, $\D_j\setminus i \subseteq \D$ be a collection of random variables in $\D_j$ except $d_i$.  An evidence potential function $\psi_j$ associated with each factor node $j \in \E$ is defined as a function of random variables $\D_j$ of its neighboring nodes $\psi_j\fof{\D_j} = \psi_j\fof{d_j}$.  Similarly, the potential function $\psi_k(\D_k)$ at the $k$th dependency factor node $k \in \S$ is defined as a function of the random variables $\D_k$ associated with its neighboring nodes.  Therefore, the factor graph represents the joint distribution of disparity labels assigned to each of the pixel locations as
    \begin{align}
        p\fof{\D} &= \frac{1}{Z} \prod_{k \in \F} \psi_k\fof{\D_k}& \label{eq:Joint dist}\\
            &=\frac{1}{Z} \prod_{j \in \E} \psi_j\fof{d_j} \prod_{k \in \S} \psi_k\fof{\D_k} & \nonumber
    \end{align}
where, $Z$ is the partitioning function.  Probability of assigning various disparity labels to each pixel $i$ can be obtained by marginalizing equation~\eqref{eq:Joint dist} with respect to $\D\setminus i$.  
    \begin{equation} \label{eq:disp marginal}
        p\fof{d_i} = \frac{1}{Z} \sum_{\D\setminus d_i}  \quad \prod_{j \in \E} \psi_j\fof{d_j} \prod_{k \in S} \psi_k\fof{\D_k}
    \end{equation}
This provides a sum-product formulation \cite{pearl1982reverend,kschischang2001factor} for determining likely disparity labels for each of the pixel locations $i$ based on \textit{a priori} disparity information and spatial dependency characteristics of disparities in a stereo image pair.

For an approximate and efficient computation of marginal beliefs or probabilities in equation~\eqref{eq:disp marginal} using \textit{loopy belief propagation}, local information available in each node is shared with neighboring nodes as variable-to-dependency factor messages $\mu_{i\in \V \rightarrow f\in \S}$ and factor-to-variable messages $\mu_{f\in \F \rightarrow i\in \V}$ until convergence \cite{barber2012bayesian,murphy2013loopy}.  Each outgoing message from a node is defined as a function of incoming messages at the given node as follows \cite{barber2012bayesian}.
    \begin{align}
        \mu_{i\in \V \rightarrow f\in \S} &= \prod_{g \in \collection{ne\fof{i}\setminus f}} \, \mu_{g \rightarrow i}\fof{d_i} & \label{eq:msg v-to-f}\\
        \mu_{f\in \F \rightarrow i\in \V} &= \sum_{\D_f\setminus d_i} \, \psi_f\fof{\D_f} \prod_{j \in \collection{ne\fof{f}\setminus i}} \, \mu_{j\rightarrow f}\fof{d_j} & \label{eq:msg f-to-v}
    \end{align}

\subsection{Probabilistic Model}
\label{sec:prob model}
For approximate inference on disparity label assignment, message structures for loopy belief propagation in equations~\eqref{eq:msg v-to-f} and \eqref{eq:msg f-to-v} and relevant potential functions were defined based on the posterior probability of disparity label assignment,
\begin{equation*}
    p\fof{d_i \mid \collection{f_j\fof{\D_j}, \forall j: i \in ne\fof{j}}} \propto p\fof{ \collection{f_j\fof{\D_j}, \forall j: i \in ne\fof{j}} \mid d_i} \, p\fof{d_i}
\end{equation*}
where, $f_j\fof{\D_j}$ is a function of the state of all variable nodes neighboring the factor node $j$.
In our proposed method, the most relevant and compact spatial dependencies are defined independently at each pixel $i$ using local image characteristics as described in Section~\ref{sec:filters}.  Therefore, the joint likelihood term can be simplified as
\begin{equation*}
    p\fof{ \collection{f_j\fof{\D_j}, \forall j: i \in ne\fof{j}} \mid d_i} = \prod_{\forall j: i \in ne\fof{j}} p\fof{f_j\fof{\D_j} \mid d_i}
\end{equation*}
and the posterior probability updated as
\begin{equation} \label{eq:posterior}
    p\fof{d_i \mid \collection{f_j\fof{\D_j}, \forall j: i \in ne\fof{j}}} \propto p\fof{d_i} \, \prod_{\forall j: i \in ne\fof{j}} p\fof{f_j\fof{\D_j} \mid d_i}
\end{equation}

It can be observed that the posterior probability of $d_i$ without conditioning on a dependency factor node $k \in ne\fof{i}$
\begin{equation} \label{eq:posterior-1}
    p\fof{d_i \mid \collection{f_j\fof{\D_j}, \forall j: i \in \collection{ne\fof{j}\setminus k}}} \propto p\fof{d_i} \, \prod_{\forall j: i \in \collection{ne\fof{j}\setminus k}} p\fof{f_j\fof{\D_j} \mid d_i}
\end{equation}
resembles the structure of the variable-to-dependency factor node message  $\mu_{i \rightarrow k}$ in equation~\eqref{eq:msg v-to-f}.  This further suggests that individual likelihood terms $p\fof{f_j\fof{\D_j} \mid d_i}$ form the dependency factor node-to-variable node message structure in equation~\eqref{eq:msg f-to-v}.

Considering the individual likelihood terms in equation~\eqref{eq:posterior-1},
\begin{align}
    p\fof{f_j\fof{\D_j} \mid d_i} &= \sum_{\D_j\setminus d_i} p\fof{f_j\fof{\D_j}, \D_j\setminus d_i \mid d_i} & \nonumber\\
        &= \sum_{\D_j\setminus d_i} p\fof{f_j\fof{\D_j} \mid \D_j} \, p\fof{\D_j\setminus d_i \mid d_i} & \nonumber\\
        &= \sum_{\D_j\setminus d_i} p\fof{f_j\fof{\D_j}} \, p\fof{\D_j\setminus d_i \mid d_i} & \nonumber
\end{align}
and assuming that random variables $\D_j$ associated with a dependency factor node $j \in \S$ are independent as in Moon \& Gunther \cite{moon2006multiple},
\begin{align}
    p\fof{f_j\fof{\D_j} \mid d_i} &= \sum_{\D_j\setminus d_i} p\fof{f_j\fof{\D_j}} \, \prod_{d_k \in \D_j\setminus d_i} \, p\fof{d_k \mid d_i} & \label{eq:likelihood}
\end{align}
In the absence of any loops between any two variable nodes $k$ and $i$ (i.e. when there is at most one common dependency factor node $s \in \S$ between any two variable nodes $k$ and $i$), the conditional probability $p\fof{d_k \mid d_i}$ can be interpreted as the posterior probability of $d_k$ conditioned on all dependency factor functions associated with the $k$th variable node, i.e. $p\fof{d_k \mid d_i} \propto p\fof{d_k \mid \collection{f_l\fof{D_l}, \forall l : l \in ne\fof{k}}}$.   By excluding the factor function $f_s\fof{\D_s}$ associated with the factor node $s \in \collection{ne\fof{k} \cap ne\fof{i}}$ in $p\fof{d_k \mid d_i}$, the individual likelihood expression in equation~\eqref{eq:likelihood} resembles the dependency factor node-to-variable node message structure $\mu_{j\rightarrow i}$ in equation~\eqref{eq:msg f-to-v}.

\subsection{Message Passing Implementation}
Based on the message-passing structures in our factor graph model, the variable-to-dependency factor node message $\mu_{i\in \V \rightarrow f\in \S}$ in equation~\eqref{eq:msg v-to-f} was approximated as the posterior probability of $d_i$ conditioned on (satisfying) all of its neighboring factor nodes except the factor node $f$ to which the message is sent as in equation~\eqref{eq:posterior-1}.  Similarly, the factor node-to-variable node messages $\mu_{f\in \F \rightarrow i\in \V}$ in equation~\eqref{eq:msg f-to-v} was approximated as the likelihood of satisfying spatial dependency among the random variable states of all variables nodes associated with the factor node $f$ except $d_i$.

It can be observed that, for evidence factor nodes $f \in \E$, the factor-to-variable messages in equation~\eqref{eq:msg f-to-v} simplifies as $\mu_{{f \in \E} \rightarrow i\in \V} = \psi_f\fof{d_i}$.  This supplies fixed prior information about the state $d_i$ of the $i$th variable node by restricting messages from variable nodes to only dependency factor nodes as in equation~\eqref{eq:msg v-to-f}.  We estimated the \textit{a priori} distribution $p\fof{d_i}$ from a \textit{disparity cost volume} containing the cost of assigning all possible disparity labels at each pixel location.  Details of building the disparity cost volume is presented in Section~\ref{sec:a priori}.  Therefore, the approximate inference obtained using our factor graph model provides updated disparities $\D$ (an optimal surface within the cost volume) based on their posterior probabilities.

The potential function $\psi_f\fof{\D_f}$ in equation~\eqref{eq:msg f-to-v} and its probabilistic representation  $p\fof{f_j\fof{\D_j}}$ in equation~\eqref{eq:likelihood} at the spatial dependency factor node $j$ was assigned a value of 1.0 when the states $\D_j \setminus i$ of the neighboring nodes are same as that of $d_i$; and was assigned a value of 0.0 when the states $\D_j \setminus i$ are different from that of $d_i$.  This enforces spatial dependencies among the neighboring pixels in the final disparity map $\D$.  In addition, this further reduces the number of marginalization operations in equations~\eqref{eq:msg f-to-v} and \eqref{eq:likelihood}.

For each stereo image pairs, each evidence factor nodes $e \in \E$ were initialized with a priori probabilities $p\fof{d_i}$ of the variable node $i \in ne\fof{e}$ as the evidence factor node-to-variable node messages.  The initial message from other variable and dependency factor nodes were set to be a uniform probability vector representing equally likely states.  After initial message passing from the evidence factor nodes $E$, message exchange continues among all graph nodes $\V \cup \F$ until message convergence.  We utilized an $L_2$ measure of 
    \begin{equation}\label{eq:convergence}
        \epsilon^{t+1} = \Vert \D^{t+1} - \D^t\Vert
    \end{equation}
for assessing message convergence at message passing iteration $t+1$.

\subsection{Disparity Cost Volume and \textit{a Priori} Disparity Distribution $p\fof{d_i}$}
\label{sec:a priori}
\begin{figure}
    \includegraphics[width=1\textwidth,keepaspectratio]{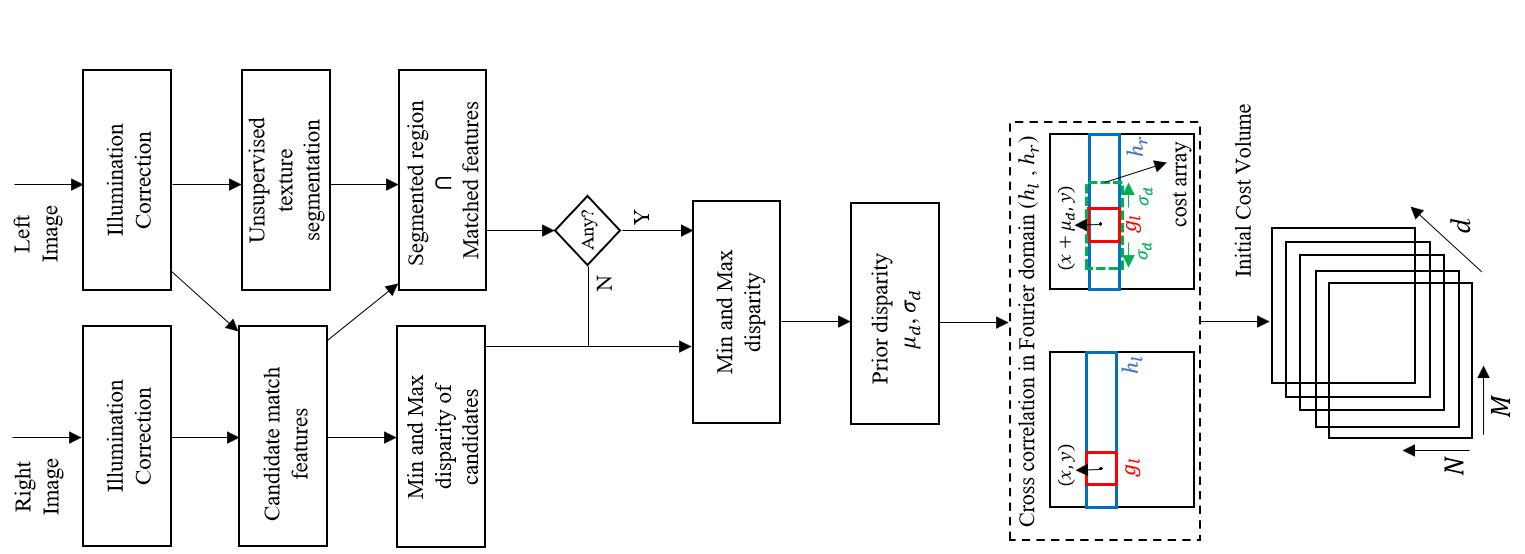}\centering
    \caption{Schematic representation of sparse cost volume calculations.}
    \label{fig:zonal disparity}       
\end{figure}

For any given stereo image pair, let, $C\fof{i, d_i}$ be a cost volume representing the cost of assigning a disparity label $d_i$ to location $i$.  Thus the final disparity map for a given stereo image pair will be an optimal surface within the cost volume $C\fof{i, d_i}$.  Fig.~\ref{fig:zonal disparity} shows a schematic representation of algorithmic steps used for disparity cost volume calculation.

For computationally efficient disparity estimation, the reference image was segmented using an unsupervised texture segmentation method \cite{jain1991unsupervised}.  In brief, sharp image segment boundaries were derived using a Gabor filter bank and image boundaries were aggregated using $k$-means clustering to generate an image segmentation map. Within each of the segmented regions, highly confident disparity estimates at several candidate locations were obtained using an eigen-based feature matching method \cite{shi1994good}.  Using the zonal / regional disparity distributions, disparity cost at each of the location $i$ was calculated using a normalized cross-correlation measure in the frequency domain.  Within each segmented region, only disparity labels ranging from $\left[\mu_d - \sigma_d, \mu_d + \sigma_d\right]$  were considered based on the distribution of the disparity labels within the segmented region as shown in Figure~\ref{fig:zonal disparity}. This results in a sparse disparity cost volume $C(i, d_i)$ and thus facilitates a faster inference due to reduced marginalization limits in equation~\eqref{eq:disp marginal}.  A detailed description of the initial cost volume computation as part of a hybrid of cross-correlation and scene segmentation (HCS) algorithm is available elsewhere \cite{Shabanian2021Hybrid}. 

\textit{A priori} probability of assigning a disparity label $d_i$ to the $i$th pixel location in the reference image was estimated from the cost volume $C\fof{i, d_i}$ as
\begin{equation}
    p\fof{d_i}=\frac{C\fof{i, d_i}}{\sum_{d_j} C\fof{i, d_j}}
\end{equation}

\subsection{Determining Variable Nodes Associated with Each Dependency Factor Node}
\label{sec:filters}
We utilized edge-preserving filter kernels, namely the guided image filters (GIF)\cite{he2010guided} and bilateral filters (BF)\cite{tomasi1998bilateral} to determine neighboring variable nodes with the highest influence on the true state of disparity at each of the $i$th pixel locations.  Neighborhood dependency information from these kernels were used to define a non-symmetric, irregular, and higher-order neighborhood system based on the fact that objects at various depths in the scene may exhibit a disparity boundary along the object boundary.  Let $\fof{x, y}$ represent the 2D coordinates of the $i$th pixel.  Highly dependent neighbors of location $(x, y)$ were selected using an $\alpha$th percentile cut-off of the kernel coefficients. Coordinates of these highly dependent neighbors were used to identify the neighboring variable nodes of each dependency factor node.

In brief, guided image filtering (GIF) is an edge-preserving smoothing algorithm. At each pixel location $(x, y)$ in the reference image $\hat{r_l}$, a guided filter kernel $W_{xy}\fof{i, j}$ with smoothness parameter $\epsilon$ was estimated as
\begin{equation}
    W_{xy}\fof{m, n} = \frac{1}{\lvert \omega_{xy} \rvert^2} \sum_{\forall \fof{k, l} \in \omega_{xy}} \bigg(1 + \frac{(\hat{r_l}\fof{m, n} - \mu_{xy})\fof{\hat{r_l}\fof{k,l} - \mu_{xy}}}{\sigma^2_{xy} + \epsilon} \bigg)
\end{equation}
where $\omega_{xy}$ is the window size used for estimating local illumination characteristics namely the mean illumination $\mu_{xy}$ and standard deviation $\sigma$ at location $\fof{x, y}$.

The Bilateral filter (BF) \cite{tomasi1998bilateral} is an edge-preserving non-linear Gaussian filter with coefficients defined as a function of spatial and intensity similarities estimated respectively using a localized domain kernel and a range kernel.  Bilateral filter kernel coefficients at location $\fof{x, y}$ is given as
\begin{equation}
    W_{xy}\fof{m, n}= \exp \bigg(-\frac{\fof{x - m}^2 + \fof{y - n}^2} {2 \sigma^2_d} - \frac{\Vert \hat{r_l}\fof{x, y} - \hat{r_l}\fof{m, n}\Vert^2}{2\sigma^2_r} \bigg)
\end{equation}
where the first exponential term represents a domain kernel as a function of pixel distance with respect to the center pixel $\fof{x, y}$ and the second term represents a range kernel as a function of regional image intensity with respect to that of the center pixel $\fof{x, y}$.  Parameters $\sigma_d$ and $\sigma_r$ controls the extent of influence neighboring pixels have on the domain and range filters respectively.


\SetNlSty{textbf}{}{:}
\SetNlSkip{1.25em}
\RestyleAlgo{ruled}
\IncMargin{1em}
\begin{algorithm}
    \SetKwData{Left}{left}\SetKwData{This}{this}\SetKwData{Up}{up}
    \SetKwFunction{Union}{Union}\SetKwFunction{FindCompress}{FindCompress}
    \SetKwInOut{Input}{Input}\SetKwInOut{Output}{Output}
    \Input{A sparse disparity cost volume $C(i,d_i)$ from a stereo pair of size $M \times N$ pixels and 
            convergence threshold $\tau$.}
    \Output{An optimal disparity $\D = \collection{d_i: d_i \in [d_{\min}, d_{\max} ] \subset \Z}_{i=1}^{MN}$.}
    \BlankLine
    Estimate \textit{a priori} probability from cost volume: $p(d_i) \leftarrow \frac{C(i, d_i)}{\sum_{d_j} C(i, d_j)}$\;
    %
    Define factor graph nodes namely variable nodes $\V$, evidence factor nodes $\E$ and dependency factor nodes $\S$ with $\F = \E \cup \S$\;
    Form factor graph by connecting each evidence factor node $j \in \E$ with each variable node $i \in \V$; and each variable node $i \in \V$ with one or more dependency factor nodes $k \in \S$ as in Sections~\ref{sec:FG structure} and \ref{sec:filters}\;
    Initialize evidence factor nodes with the prior probabilities: $j \in \E \leftarrow  p(d_i)$, where $ i \in ne(j)$\;    
    Associate each variable node $i$ with a random variable $d_i$ to represent disparity label at the $i$th pixel location\;
    Define an evidence potential function $\psi_j, j \in E$ as a function of random variables $\D_j$ of its neighboring nodes $\psi_j(\D_j)=\psi_j(d_i)$\;
    Define a dependency potential function $\psi_k, k \in \S$ as a function of random variables $\D_k$ of its neighboring nodes $\psi_k(\D_k)$\;
    \While(Loopy belief propagation){convergence error $\epsilon^{t+1}$ in \eqref{eq:convergence} $>$ threshold $\tau$}{
    \tab Send variable-to-dependency-factor messages: $\mu_{i\in \V \rightarrow f\in \S }$ as in \eqref{eq:msg v-to-f}\;
    \tab Send factor-to-variable messages: $\mu_{f\in \F \rightarrow i\in \V}$ as in \eqref{eq:msg f-to-v}\;
    }
    Determine \textit{maximum a posteriori} disparity estimate $\hat{d}_i$ at each pixel as in equation~\eqref{eq:posterior}: $\hat{d}_i = \argmax_{d_i} p\fof{d_i \mid \collection{f_j\fof{\D_j}, \forall j: i \in ne\fof{j}}}$\;
    Update disparity estimates in occluded regions using \textit{left-right consistency check} as in Section~\ref{sec:postprocessing}\;
    \caption{Probabilistic factor graph-based Disparity Estimation (FGS) Algorithm}\label{algo_disjdecomp}
    \label{alg:fgs}
\end{algorithm}\DecMargin{1em}
\subsection{Disparity Estimates}
Upon message passing convergence, an approximate estimate of the posterior probability of assigning a disparity label $d_i$ is available in each variable node as given in equation~\eqref{eq:posterior}.  A \textit{maximum a posterior} disparity estimate was determined at each pixel location $i$ based on the disparity label $d_i$ with the maximum posterior probability at the respective pixel $i$.
\subsection{Post-processing the Disparity Maps}
\label{sec:postprocessing}
Occluded pixels were identified based on a lack of consistency or agreement between the disparity maps estimated using the left-right order vs right-left order of each stereo pair \cite{cochran19923}. For each occluded pixel, the disparity estimate from its nearest non-occluded pixel within the same scanline (row) was assigned.  Further, a weighted median filter was used to minimize spurious disparity assignments in the occluded region \cite{brownrigg1984weighted}. 
%
\section{Experimental Results and Discussion}
\label{Experiments}

Algorithmic steps of the proposed factor graph method are presented in Algorithm~\ref{alg:fgs}.  We evaluated the proposed method using stereo images from the Middlebury benchmark stereo datasets \cite{scharstein2003high}, \cite{scharstein2007learning}, \cite{hirschmuller2007evaluation}, and \cite{scharstein2014high}.  We present a detailed evaluation of the proposed FGS algorithm followed by a comprehensive comparison of its performance with other state-of-the-art disparity estimation algorithms using Middlebury evaluation dataset version 3.0 \cite{scharstein2002taxonomy}.
\subsection{FGS Parameters and FGS Implementation}
The FGS algorithm was implemented in MATLAB 2018b and its performance was evaluated using an Intel(R) Xeon(R) workstation with E3-1271 v3, 3.6 GHz processor. 

For illumination correction, high pass filters of size $21 \times 21$ pixels were obtained from a low-pass averaging filter.  For unsupervised texture segmentation of the reference image, Gabor filters were designed \cite{jain1991unsupervised} with filter orientations between $0\degree-150\degree$ degrees in steps of $30\degree$, and wavelength starting from $2.83$ up to the magnitude of hypotenuse of the input image. $K$-means clustering algorithm was initialized with $K=15$ with $5$ replicates and ran for a maximum of $500$ iterations. For cost volume calculations, templates of size $3 \times 3$ pixels were used.  For identifying variable nodes connected to each dependency factor node, bilateral filter kernel of size $7 \times 7$ pixels, domain kernel parameter of $\sigma_d = 3$, range kernel parameter of $\sigma_r = 0.1$ and coefficient percentile cut-off of $\alpha=97$ were used.  We observed that the smallest kernel sizes along with a higher percentile cut-off $\alpha$ identified fewer but highly dependent neighboring nodes.  Therefore, the computational cost of message passing was significantly reduced with fewer but highly reliable neighboring variable nodes connected to each dependency factor node.  
\subsection{Performance Metrics}
For performance evaluation, we used the common performance metrics available for assessing the accuracy of the estimated disparity maps namely, the disparity error maps, peak signal-to-noise ratio (PSNR), and average absolute error (Avg. err in pixels). Disparity error maps were computed as location-wise difference between the estimated disparity  $\hat{d}(x,y)$ and its ground-truth $d(x,y)$ as $\hat{d}(x,y) - d(x,y)$.  PSNR provides a measure of similarity between an estimated disparity map $\hat{d}(x,y)$ of size $M \times N$ pixels and the ground-truth disparity map $d(x,y)$ as follows.
    \begin{equation}
        PSNR=10 \log _{10}\frac{255^2 \times M \times N}{\sum_{\forall(x,y)}\fof{\hat{d}(x,y)-d(x,y)}^2}
    \end{equation}
A thresholded average disparity error metric with a disparity threshold of $T$ pixels was defined as  
    \begin{equation}
         Bad=\fof{\frac{1}{MN}\sum_{\forall(x,y)}(|\hat{d}(x,y)- d(x,y)|>T)}\times 100
    \end{equation}
Average disparity errors were assessed at two disparity threshold levels of $T=2$ pixels (Bad2.0) and  $T=0$ pixels (Avg. err).

In the majority of the stereo pairs tested, the FGS algorithm converged between 25 and 30 iterations based on the convergence measure in equation~\eqref{eq:convergence}.

\subsection{Filter Selection for Identifying FGS Variable Node Neighbors}
\begin{figure}[!ht]
\centering
     \begin{subfigure}[a]{0.24\textwidth}
         \includegraphics[width=\textwidth]{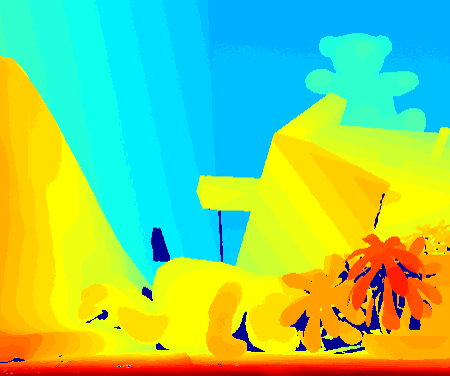}
         \caption{}
         \label{fig:3.a}
     \end{subfigure}
    \qquad
     \begin{subfigure}[a]{0.24\textwidth}
         \includegraphics[width=\textwidth]{Teddy_Initial.png}
         \caption{}
         \label{fig:3.b}
     \end{subfigure}\\
     \begin{subfigure}[a]{0.24\textwidth}
         \includegraphics[width=\textwidth]{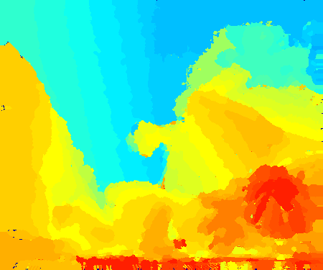}
         \caption{}
         \label{fig:3.c}
     \end{subfigure} 
     \hfill
      \begin{subfigure}[a]{0.24\textwidth}
         \includegraphics[width=\textwidth]{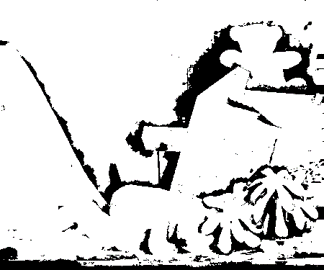}
         \caption{}
         \label{fig:3.d}
     \end{subfigure}
     \hfill
      \begin{subfigure}[a]{0.24\textwidth}
         \includegraphics[width=\textwidth]{Teddy_Bilateral.png}
         \caption{}
         \label{fig:3.e}
     \end{subfigure}
     \begin{subfigure}[a]{0.24\textwidth}
         \includegraphics[width=\textwidth]{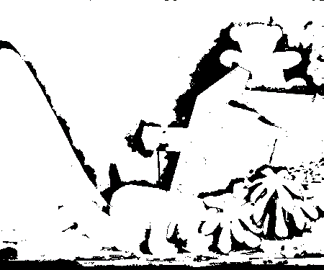}
         \caption{}
         \label{fig:3.f}
     \end{subfigure}
    \caption{Effect of the edge-preserving filters on the accuracy of the disparity map for the ``Teddy'' stereo pair estimated using the FGS algorithm. Disparity estimates in the occluded regions are not excluded in the final disparity map. (a) Ground truth, (b) Initial disparity map, (c) Disparity map estimated using the FGS algorithm with guided filters, (d) Disparity error map for Fig.~\ref{fig:3.c}, (e) Disparity map estimated using the FGS algorithm with bilateral filters, and (f) Disparity error map for Fig.~\ref{fig:3.e}.} 
    \label{fig:3}       
\end{figure}

The edge-preserving filters (Sec~\ref{sec:filters}) with the highest accuracy and computational speed were selected for identifying neighboring variable nodes for each of the dependency factor nodes in the FGS algorithm.  Figures~\ref{fig:3.c} and \ref{fig:3.e} show the disparity maps for the ``Teddy" stereo pair estimated using guided image filters and bilateral filters respectively.

Quantitative performance measures of the edge-preserving filters are presented in Table~\ref{tab:1}.  The accuracy of the FGS algorithm with guided filters (based on the Avg. err, PSNR, and Bad2.0 metrics) were slightly better than the bilateral filters.  Because the guided filters resulted in a larger number of neighboring variable nodes, the run time of the FGS algorithm, however, was higher with the guided filters when compared to the FGS algorithm with bilateral filters. Therefore, for further evaluation of the FGS algorithm, bilateral filtering was chosen as the optimal choice for identifying neighboring variable nodes in the FGS factor graph.

\begin{table}[!ht]
    \centering
    \caption{Performance of the edge-preserving filters used for identifying  variables nodes neighboring each dependency factor node in the FGS factor graph.}
    \label{tab:1}       
    \begin{tabular}{cccc}
        \hline\noalign{\smallskip}
        Algorithm & Avg. err & PSNR (dB) & Bad 2.0 (\%)\\
        \noalign{\smallskip}\hline\noalign{\smallskip}
        Initial disparity using HCS & 5.18 & 26.33 & 24.21\\
        FGS with guided filtering & \textbf{2.59} & \textbf{32.04} & \textbf{14.12} \\
        FGS with bilateral filtering & 2.60 & 32.02 & 14.17\\
        \noalign{\smallskip}\hline
    \end{tabular}
\end{table}
\subsection{Detailed Assessment of the FGS Algorithm using Selected Stereo Pairs}
For detailed quantitative and qualitative assessment of the FGS algorithm, we utilized stereo image pairs with differing textures, illuminations, and exposure characteristics from the Middlebury stereo dataset including 2003 \cite{scharstein2003high}, 2005 \cite{scharstein2007learning}, 2006 \cite{hirschmuller2007evaluation}, and 2014 \cite{scharstein2014high} stereo datasets. The stereo pairs selected for assessment were the \textit{Teddy and Cones} stereo pair (2003), \textit{Dolls stereo pair} (2005), \textit{Rocks1} stereo pair (2006), and the \textit{Motorcycle} stereo pair (2014). Assessment results based on the estimated disparity maps with and without post-processing are presented in the following sections.
\subsubsection{Assessment Results Without Post-processing:}
\begin{figure}[!ht]
\centering
   \rotatebox[origin=c]{90}{\tiny Ground truth}
     \begin{subfigure}[a]{0.19\textwidth}
        \centering
         \includegraphics[width=\textwidth]{Teddy_GT.png}
         \label{fig:4.a.1}
     \end{subfigure}
     \hfill
     \begin{subfigure}[a]{0.19\textwidth}
         \centering
         \includegraphics[width=\textwidth]{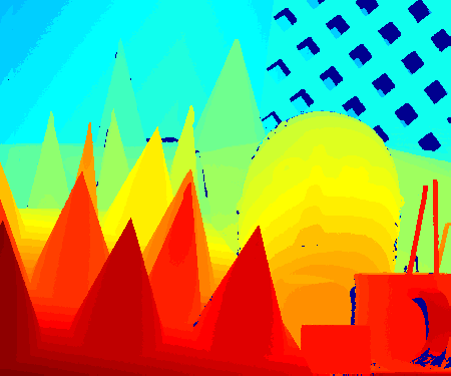}
         \label{fig:4.b.1}
     \end{subfigure} 
     \hfill
      \begin{subfigure}[a]{0.19\textwidth}
         \centering
         \includegraphics[width=\textwidth]{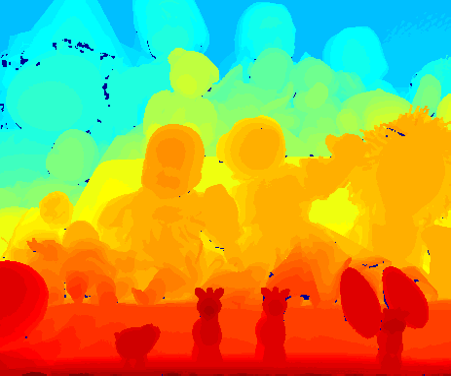}
         \label{fig:4.c.1}
     \end{subfigure}
      \hfill
      \begin{subfigure}[a]{0.19\textwidth}
         \includegraphics[width=\textwidth]{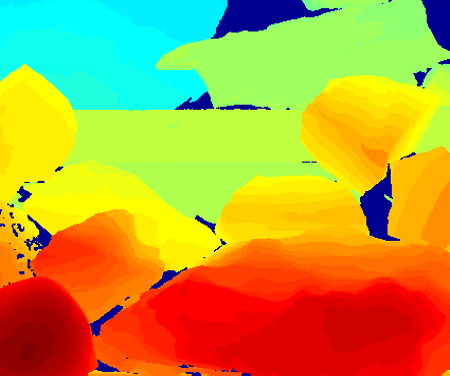}
         \label{fig:4.d.1}
     \end{subfigure}     
     \hfill
      \begin{subfigure}[a]{0.19\textwidth}
         \includegraphics[width=\textwidth]{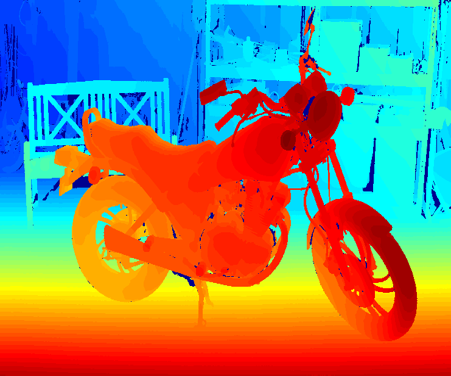}
         \label{fig:4.e.1}
     \end{subfigure}  \\   
     \rotatebox[origin=c]{90}{\tiny Initial Disparity}
     \begin{subfigure}[a]{0.19\textwidth}
         \centering
         \includegraphics[width=\textwidth]{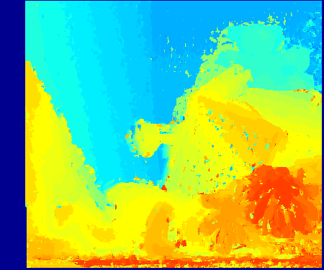}
         \label{fig:4.a.2}
     \end{subfigure}
     \hfill
     \begin{subfigure}[a]{0.19\textwidth}
         \centering
         \includegraphics[width=\textwidth]{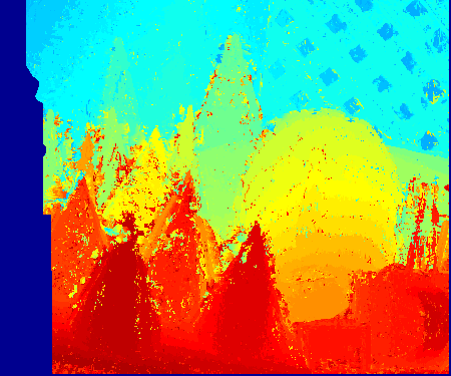}
         \label{fig:4.b.2}
     \end{subfigure} 
     \hfill
      \begin{subfigure}[a]{0.19\textwidth}
         \centering
         \includegraphics[width=\textwidth]{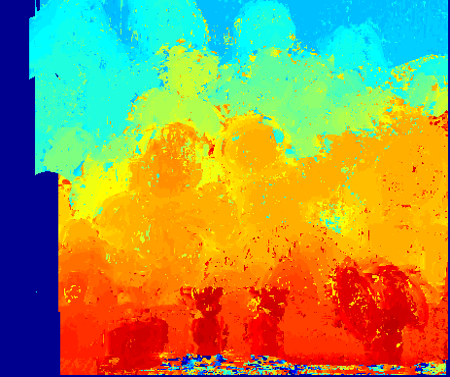}
         \label{fig:4.c.2}
     \end{subfigure}
      \hfill
      \begin{subfigure}[a]{0.19\textwidth}
         \includegraphics[width=\textwidth]{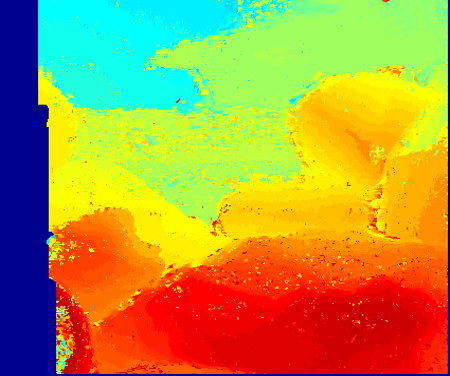}
         \label{fig:4.d.2}
     \end{subfigure}     
     \hfill
      \begin{subfigure}[a]{0.19\textwidth}
         \includegraphics[width=\textwidth]{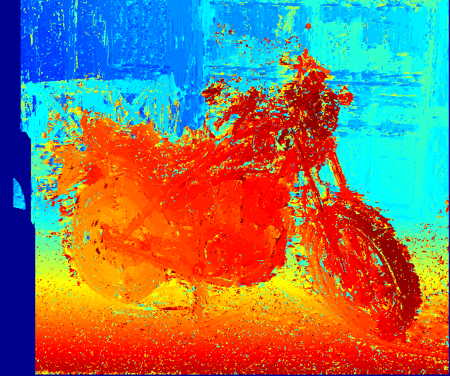}
         \label{fig:4.e.2}
     \end{subfigure}  \\   
   \centering \rotatebox[origin=c]{90}{\tiny FGS disparity}
      \begin{subfigure}[a]{0.19\textwidth}
         \centering
         \includegraphics[width=\textwidth]{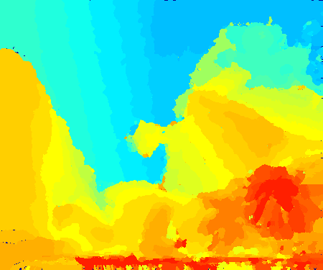}
         \label{fig:4.a.3}
     \end{subfigure}
     \hfill
     \begin{subfigure}[a]{0.19\textwidth}
         \centering
         \includegraphics[width=\textwidth]{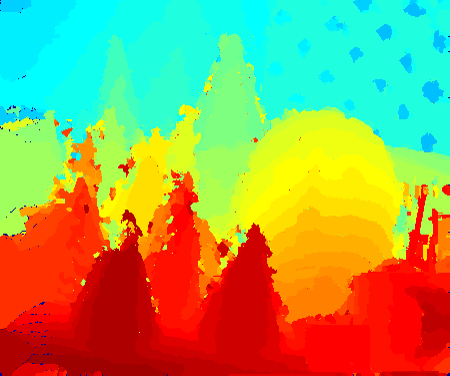}
         \label{fig:4.b.3}
     \end{subfigure} 
     \hfill
      \begin{subfigure}[a]{0.19\textwidth}
         \centering
         \includegraphics[width=\textwidth]{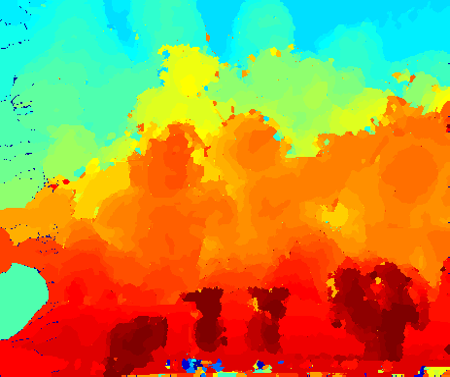}
         \label{fig:4.c.3}
     \end{subfigure}
      \hfill
      \begin{subfigure}[a]{0.19\textwidth}
         \includegraphics[width=\textwidth]{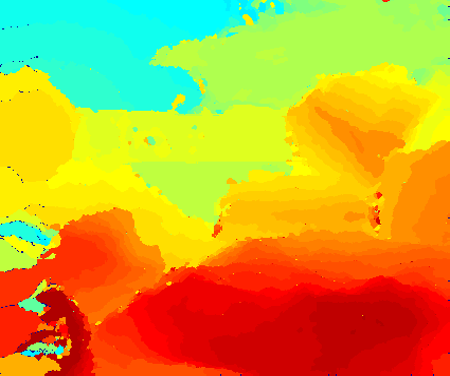}
         \label{fig:4.d.3}
     \end{subfigure}     
     \hfill
      \begin{subfigure}[a]{0.19\textwidth}
         \includegraphics[width=\textwidth]{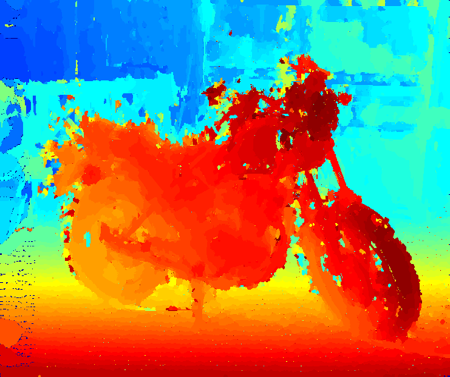}
         \label{fig:4.e.3}
     \end{subfigure}  \\   
      
      \rotatebox[origin=c]{90}{\tiny \tab Disparity error map}
      \begin{subfigure}[a]{0.19\textwidth}
         \centering
         \includegraphics[width=\textwidth]{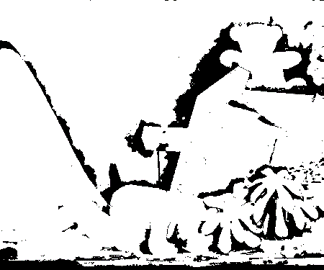}
         \label{fig:4.a.4} \caption{}
     \end{subfigure}
     \hfill
     \begin{subfigure}[a]{0.19\textwidth}
         \centering
         \includegraphics[width=\textwidth]{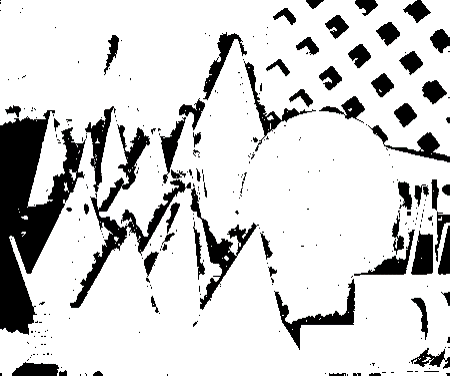}
         \label{fig:4.b.4} \caption{}
     \end{subfigure} 
     \hfill
      \begin{subfigure}[a]{0.19\textwidth}
         \centering
         \includegraphics[width=\textwidth]{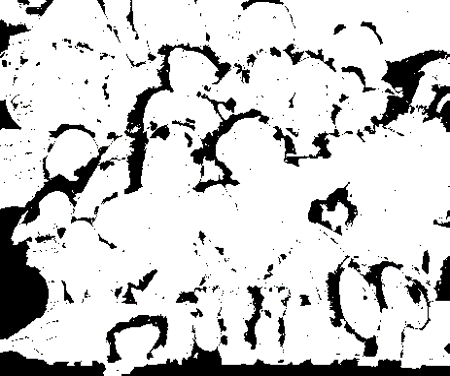}
         \label{fig:4.c.4} \caption{}
     \end{subfigure}
      \hfill
      \begin{subfigure}[a]{0.19\textwidth}
         \includegraphics[width=\textwidth]{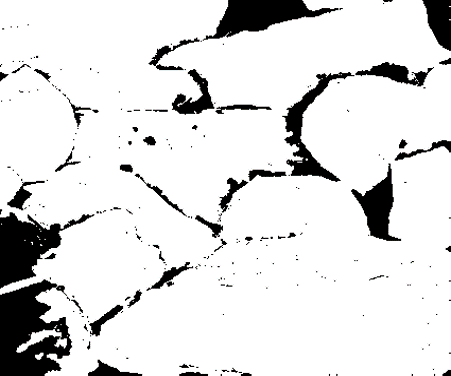}
         \label{fig:4.d.4} \caption{}
     \end{subfigure}     
     \hfill
      \begin{subfigure}[a]{0.19\textwidth}
         \includegraphics[width=\textwidth]{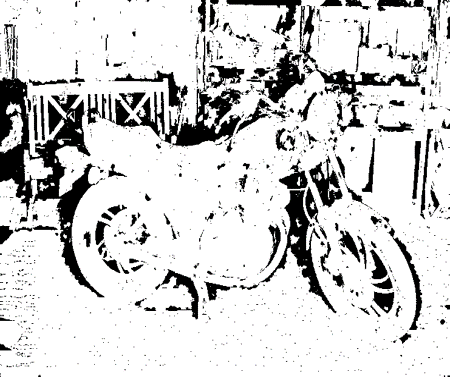}
         \label{fig:4.e.4} \caption{}
     \end{subfigure}  \\       
    \caption{Estimated disparity maps and error maps for selected Middlebury stereo pairs without post-processing. (a) Teddy (dataset 2003), (b) Cones (dataset 2003), (c) Dolls (dataset 2005), (d) Rocks1 (dataset 2006), and (e) Motorcycle (dataset 2014). Disparity estimates in the occluded regions were not excluded in the final disparity maps.}
    \label{fig:4}       
\end{figure}

For the selected stereo pairs, disparity maps estimated by the HCS algorithm without cost aggregation and post-processing, disparity maps estimated by the FGS algorithm without any post-processing and the disparity errors for the FGS algorithm are shown in Fig.~\ref{fig:4}.  A summary of the assessment metrics without post-processing the disparity maps is presented in Table~\ref{tab:2}.  It can be observed that the FGS algorithm significantly improved over the initial disparity estimates generated by the HCS algorithm without cost aggregation.  In addition, the FGS algorithm is able to identify disparities in occluded regions.
\begin{table}[!ht]
    \centering
    \caption{Performance measures of the FGS algorithm for selected Middlebury stereo pairs without post-processing the disparity maps. Disparity estimates in the occluded regions were not excluded in the final disparity maps.  FGS estimates significantly improved over the initial disparity estimates by all performance metrics.}
    \label{tab:2}       
    \begin{tabular}{cccccccc}
    \hline\noalign{\smallskip}
        Images & \multicolumn{3}{c}{Initial} & & \multicolumn{3}{c}{FGS Algorithm}\\
        \cline{2-4}\cline{6-8}\noalign{\smallskip} 
        &Avg.err & PSNR(dB) & Bad2.0(\%) & & Avg.err & PSNR(dB) & Bad2.0(\%)\\
        \noalign{\smallskip}\hline\noalign{\smallskip}
        Teddy      & 5.18 & 26.33 & 24.21 & & 2.60 & 32.02 & 14.17\\
        Cones      & 5.88 & 25.20 & 25.69 & & 2.78 & 32.35 & 17.60\\
        Dolls      & 7.51 & 23.51 & 29.90 & & 3.20 & 30.75 & 22.47\\
        Rocks1     & 6.35 & 24.42 & 21.97 & & 3.29 & 30.15 & 13.58\\ 
        Motorcycle & 5.96 & 26.07 & 33.42 & & 3.81 & 29.45 & 20.04\\
    \noalign{\smallskip}\hline\noalign{\smallskip}
    \end{tabular}
\end{table}
%
\subsubsection{Assessment Results with Post-processing} 
\begin{figure}[!ht]
\centering
     \rotatebox[origin=c]{90}{\tiny Ground truth}
     \begin{subfigure}[a]{0.19\textwidth}
         \centering
         \includegraphics[width=\textwidth]{Teddy_GT.png}
         \label{fig:5.a.1}
     \end{subfigure}
     \hfill
     \begin{subfigure}[a]{0.19\textwidth}
         \centering
         \includegraphics[width=\textwidth]{Cones_GT.png}
         \label{fig:5.b.1}
     \end{subfigure} 
     \hfill
      \begin{subfigure}[a]{0.19\textwidth}
         \centering
         \includegraphics[width=\textwidth]{Dolls_GT.png}
         \label{fig:5.c.1}
     \end{subfigure}
      \hfill
      \begin{subfigure}[a]{0.19\textwidth}
         \includegraphics[width=\textwidth]{Rocks1_GT.png}
         \label{fig:5.d.1}
     \end{subfigure}     
     \hfill
      \begin{subfigure}[a]{0.19\textwidth}
         \includegraphics[width=\textwidth]{Motorcycle_GT.png}
         \label{fig:5.e.1}
     \end{subfigure}  \\ 
         \rotatebox[origin=c]{90}{\tiny \hspace{0.5cm} FGS final disparity}
     \begin{subfigure}[a]{0.19\textwidth}
         \centering
         \includegraphics[width=\textwidth]{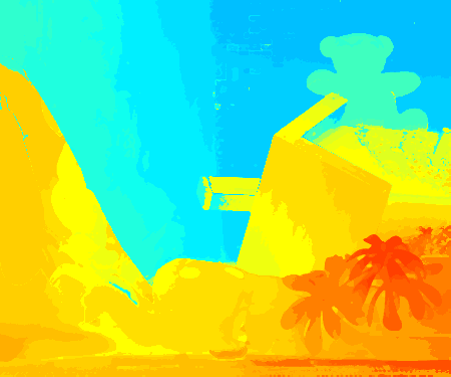}
         \label{fig:5.a.2}
     \end{subfigure}
     \hfill
     \begin{subfigure}[a]{0.19\textwidth}
         \centering
         \includegraphics[width=\textwidth]{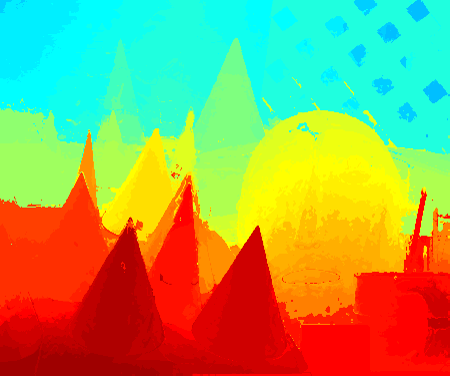}
         \label{fig:5.b.2}
     \end{subfigure} 
     \hfill
      \begin{subfigure}[a]{0.19\textwidth}
         \centering
         \includegraphics[width=\textwidth]{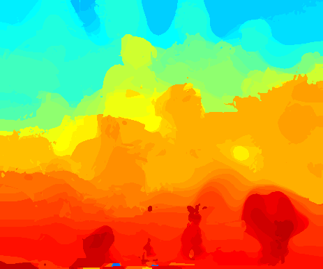}
         \label{fig:5.c.2}
     \end{subfigure}
      \hfill
      \begin{subfigure}[a]{0.19\textwidth}
         \includegraphics[width=\textwidth]{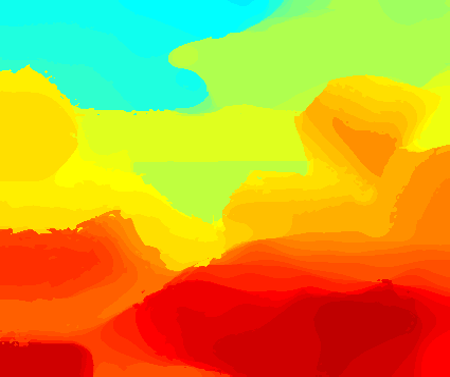}
         \label{fig:5.d.2}
     \end{subfigure}     
     \hfill
      \begin{subfigure}[a]{0.19\textwidth}
         \includegraphics[width=\textwidth]{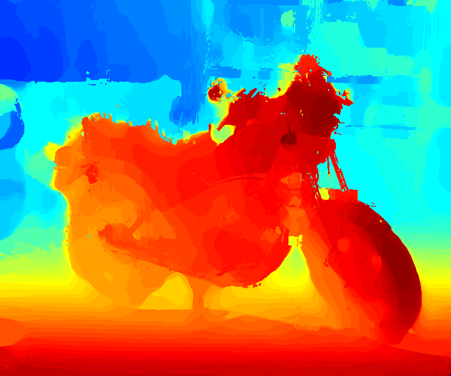}
         \label{fig:5.e.2}
     \end{subfigure}  \\   
      \rotatebox[origin=c]{90}{\tiny \tab Disparity error map}        
    \begin{subfigure}[a]{0.19\textwidth}
         \centering
         \includegraphics[width=\textwidth]{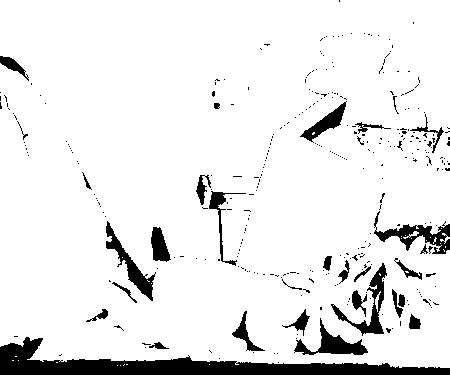}
         \label{fig:5.a.4} \caption{}
     \end{subfigure}
     \hfill
     \begin{subfigure}[a]{0.19\textwidth}
         \centering
         \includegraphics[width=\textwidth]{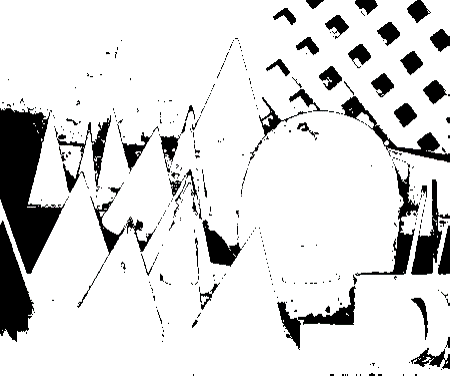}
         \label{fig:5.b.4} \caption{}
     \end{subfigure} 
     \hfill
      \begin{subfigure}[a]{0.19\textwidth}
         \centering
         \includegraphics[width=\textwidth]{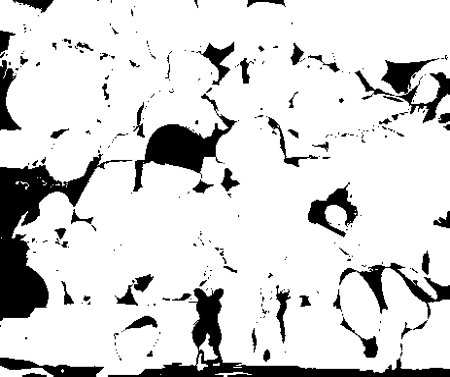}
         \label{fig:5.c.4} \caption{}
     \end{subfigure}
      \hfill
      \begin{subfigure}[a]{0.19\textwidth}
         \includegraphics[width=\textwidth]{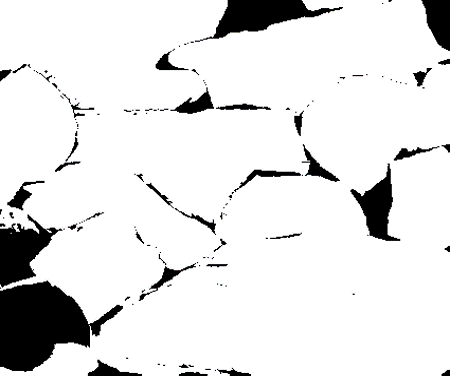}
         \label{fig:5.d.4} \caption{}
     \end{subfigure}     
     \hfill
      \begin{subfigure}[a]{0.19\textwidth}
         \includegraphics[width=\textwidth]{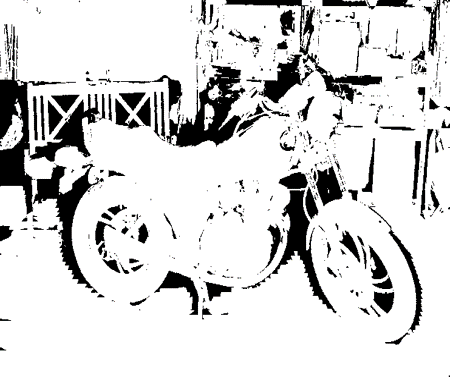}
         \label{fig:5.e.4} \caption{}
     \end{subfigure}  \\ 
    \caption{Disparity maps after post-processing and corresponding error maps for selected Middlebury stereo pairs. (a) Teddy (dataset 2003), (b) Cones (dataset 2003), (c) Dolls (dataset 2005), (d) Rocks1 (dataset 2006), and (e) Motorcycle (dataset 2014). Disparity estimates in the occluded regions were not excluded in the final disparity maps.}
    \label{fig:5}       
\end{figure}

Disparity estimates with post-processing and disparity error maps for selected stereo pairs are shown in Fig.~\ref{fig:5}.  A summary of the assessment metrics after post-processing the disparity estimates is presented in Table~\ref{tab:3}.

\begin{table}[!ht]
    \centering
    \caption{Performance measures of the FGS algorithm for selected Middlebury stereo pairs after post-processing. Disparity estimates in the occluded regions were not excluded.  FGS estimates significantly improved over the initial disparity estimates by all performance metrics.}
    \label{tab:3}       
    \begin{tabular}{cccccccc}
    \hline\noalign{\smallskip}
        Images & \multicolumn{3}{c}{FGS-based} & & \multicolumn{3}{c}{Final}\\
        \cline{2-4}\cline{6-8}\noalign{\smallskip} 
        &Avg.err & PSNR(dB) & Bad2.0(\%) & & Avg.err & PSNR(dB) & Bad2.0(\%)\\
        \noalign{\smallskip}\hline\noalign{\smallskip}
        Teddy      &  2.60 & 32.02 & 14.17 & & 1.90 & 33.25 & 9.55\\
        Cones      &  2.78 & 32.35 & 17.60 & & 2.32 & 33.33 & 15.11\\
        Dolls      &  3.20 & 30.75 & 22.47 & & 2.04 & 32.52 & 18.98\\
        Rocks1     &  3.29 & 30.15 & 13.58 & & 2.78 & 31.10 & 12.06\\ 
        Motorcycle &  3.81 & 29.45 & 20.04 & & 3.36 & 30.02 & 18.87\\
    \noalign{\smallskip}\hline\noalign{\smallskip}
    \end{tabular}
\end{table}
\subsection{Performance of the FGS Algorithm vs State-of-the-art Algorithms}
To the best of our knowledge, the proposed FGS algorithm is the first disparity estimation technique based on factor graphs.  The performance of the non-learning-based FGS algorithm was compared with recent state-of-the-art learning-based and non-learning-based disparity estimation algorithms.  All disparity estimation algorithms were evaluated using stereo pairs from the current Middlebury evaluation dataset version 3.0. In addition to the performance metrics of \textit{Avg. err}, \textit{PSNR} and \textit{Bad 2.0 \%}, a weighted average measure was calculated for each performance metric based on the level of difficulty of estimating the disparity map for each stereo pair in the Middlebury evaluation dataset version 3.0. A summary of all the assessment metrics and a weighted performance measure for the FGS algorithm is presented in Table \ref{tab:4}. 
\begin{table}[!ht]
\centering
\caption{All performance measures of the proposed FGS algorithm for the stereo pairs in the current Middlebury evaluation dataset version 3.0. Disparity estimates in the occluded regions were not excluded in the final disparity maps.}
\label{tab:4}       
\begin{tabular}{cccccccc}
\hline\noalign{\smallskip}
    Images & \multicolumn{3}{c}{FGS-based}\\
    \cline{2-4} 
    \noalign{\smallskip}(Weight) &Avg.err & PSNR(dB) & Bad2.0(\%) \\
    \noalign{\smallskip}\hline\noalign{\smallskip}
    Adiron (8)	&03.63	&30.10	&17.96\\
    ArtL (8)	&04.81	&30.70	&44.81\\
    Jadepl (8)	&26.72	&18.33	&47.01\\
    Motor (8)	&03.36	&30.02	&18.87\\
    MotorE (8)	&05.90	&27.11	&28.48\\
    Piano (8)	&04.23	&30.30	&24.12\\
    PianoL (4)	&07.10	&27.11	&35.21\\
    Pipes (8)	&07.71	&25.71	&39.25\\
    Playrm (4)	&05.30	&27.10	&30.11\\
    Playt (4)	&03.19	&31.98	&28.10\\
    PlaytP (8)	&03.05	&31.95	&24.36\\
    Recyc (8)	&03.71	&31.59	&32.54\\
    Shelvs (4)	&05.30	&27.81	&34.07\\
    Teddy (8)	&01.90	&33.25	&09.55\\
    Vintage (4)	&10.22	&29.11	&54.10\\ 
    \hline
    \textbf{Weighted}\\
    \textbf{Average} & \textbf{6.45} &\textbf{28.85}&\textbf{30.22}\\
\noalign{\smallskip}\hline\noalign{\smallskip}
\end{tabular}
\end{table}
\subsection{FGS vs Non-learning based Disparity Estimation Methods}
Performance of the FGS algorithm was compared with 12 recently developed non-learning-based disparity estimation procedures including 7 local methods, 3 global methods, and one fusion method using both local and global approaches for disparity estimation.  The local methods were: weighted adaptive cross-region-based guided image filtering method (ACR-GIF-OW) \cite{kong2021local}, real-time stereo matching algorithm with FPGA architecture (MANE) \cite{vazquez2021real}, adaptive support-weight approach in pyramid structure (DAWA-F) \cite{navarro2019semi}, encoding-based approaches PPEP-GF \cite{fu2021pixel}, absolute difference (AD) and census transform-based stereo matching with guided image filtering (ADSR-GIF) \cite{kong-ADSR-GIF}, the sum of absolute difference (SAD) based stereo matching aggregated with adaptive weighted bilateral filter (SM-AWP) \cite{razak2019effect}, statistical \textit{maximum a posteriori} estimation of MRF disparity labels (SRM) \cite{okae2020robust}. The global disparity estimation procedures chosen for comparison were: binocular narrow-baseline stereo matching procedure using a max-tree data structure (MTS) \cite{brandt2020efficient} and its improvement (MTS-2) \cite{brandt2020mtstereo}, and an accelerated multi-block matching (MBM) algorithm on GPU \cite{chang2018real}. FASW  approach\cite{wu2019stereo} uses both local and global strategies and is based on a census transform with adaptive support weight.
\begin{table}[!ht]
\renewcommand{\tabcolsep}{2pt}
\centering
\caption{Performance (Avg. err) of the FGS algorithm vs state-of-the-art \textit{non-learning-based disparity estimation algorithms} for stereo pairs in the Middlebury evaluation dataset version 3.0.  The weight of a stereo pair represents the level of difficulty in estimating disparity map from the respective stereo pair.  The algorithm with the highest performance is highlighted in bold.}
\label{tab:5}   
\begin{tabular}{cccccccccccccc}
\hline\noalign{\smallskip}
   \rot{90}{Images (Weight)} & \rot{90}{FGS} & \rot{90}{HCS \cite{Shabanian2021Hybrid}} & \rot{90}{ACR-GIF-OW \cite{kong2021local}} & \rot{90}{MANE \cite{vazquez2021real}} & \rot{90}{SRM \cite{okae2020robust}} & \rot{90}{MTS \cite{brandt2020efficient}} & \rot{90}{DAWA-F \cite{navarro2019semi}} & \rot{90}{PPEP-GF \cite{fu2021pixel}} & \rot{90}{MTS-2 \cite{brandt2020mtstereo}} & \rot{90}{ADSR-GIF \cite{kong-ADSR-GIF}} & \rot{90}{FASW \cite{wu2019stereo}} & \rot{90}{SM-AWP \cite{razak2019effect}} & \rot{90}{MBM \cite{chang2018real}}\\
    \noalign{\smallskip}\hline\noalign{\smallskip}
Adiron $(8)$  &3.63	&3.98	&4.53	&11.60	&2.88	&19.00	&4.37	&8.12	&21.50	&6.40	 &\textbf{2.86}	&10.5	&4.39\\
ArtL $(8)$   &4.81	&\textbf{4.31}	&8.41	&22.90	&5.96	&22.50	&13.00	&14.80	&22.40	&9.00	 &8.03	&19.9	&8.80\\
Jadepl $(8)$ &26.72	&27.22	&\textbf{22.10}	&45.90	&24.70	&123.00	&44.40	&46.90	&108.00	&26.10	 &34.70	&62.7	&37.60\\
Motor $(8)$  &\textbf{3.36}	&3.91	&7.93	&12.40	&4.46	&17.50	&7.29	&7.99	&15.30	&8.11	&5.44	&11:00	&5.76 \\
MotorE $(8)$ &5.90	&6.12	&7.88	&12.30	&\textbf{4.43}	&20.70	&7.04	&7.62	&30.60	&11.40	&4.43	&12.5	&5.56\\
Piano $(8)$ &4.23	&5.23	&6.36	&15.10	&5.73	&13.00	&\textbf{3.27}	&9.76	&10.00	&6.15	&5.54	&9.08	&6.67\\
PianoL $(4)$ & \textbf{7.10}	&7.44	&27.70	&24.70	&7.99	&32.00	&21.70	&18.80	&26.20	&34.00	&10.80	&29.7	&12.40\\
Pipes $(8)$ &7.71	&7.52	&11.00	&22.30	&\textbf{6.96}	&29.40	&15.90	&17.30	&24.60	&14.90	&10.80	&21.11	&11.80\\
Playrm $(4)$ &\textbf{5.30}	&6.16	&8.51	&31.10	&10.70	&26.90	&8.86	&19.30	&23.30	&10.50	&7.31	&20.7	&12.90\\
Playt $(4)$ &\textbf{3.19}	&3.57	&16.10	&39.90	&4.48	&27.40	&6.39	&45.50	&12.70	&16.70	&14.50	&9.5	&12.00\\
PlaytP $(8)$ &\textbf{3.05}	&3.17	&6.60	&17.30	&3.32	&12.00	&3.34	&24.50	&9.29	&10.00	&3.32	&9.75	&6.37\\
Recyc $(8)$ &3.71	&3.62	&4.26	&9.67	&2.92	&17.50	&3.89	&7.64	&11.00	&4.20	&\textbf{2.84}	&7.18	&3.67\\
Shelvs $(4)$ &\textbf{5.30}	&5.51	&13.10	&22.50	&7.41	&12.10	&11.10	&17.20	&11.80	&9.97	&8.70	&11.4	&11.80\\
Teddy $(8)$ &\textbf{1.90}	&2.10	&2.86	&12.50	&1.92	&8.11	&3.39	&7.11	&6.67	&3.35	&2.83	&9.44	&3.74\\
Vintage $(4)$ &10.22	&10.62	&7.77	&51.00	&15.80	&27.20	&\textbf{6.48}	&23.40	&33.80	&10.90	&6.79	&16.8	&14.10\\
\hline
\textbf{Weighted}\\
\textbf{Average} &\textbf{6.45} &6.71	&9.48	&21.33	&6.92	&27.64	&10.65	&17.11	&25.06	&11.25	&8.39	&17.38	&10.08\\
\noalign{\smallskip}\hline\noalign{\smallskip}
\end{tabular}
\end{table}

Avg. err metric and a weighted average of Avg. err for the FGS algorithm and the non-learning-based disparity estimation procedures are presented in Table \ref{tab:5}.  Image weight given in Table \ref{tab:5} represents the level of difficulty in estimating the disparity from a given stereo pair.  In general, the proposed FGS algorithm provided higher accuracy for all the stereo pairs comparable to that of other non-learning-based methods.  The FGS method provided the lowest weighted average of Avg. err of 6.45 pixels among all the non-learning-based methods.  Further, the FGS method provided the lowest estimation error (Avg. err) for 3 out of the 10 difficult stereo pairs (image weight = 8) and for 4 out of the 5 moderately difficult stereo pairs (image weight = 4). 
\subsection{FGS vs Learning-based Disparity Estimation Methods}
Performance of the FGS algorithm was also compared with the following recently developed learning-based disparity procedures: a method based on a fusion of convolutional neural networks (CNN) and conditional random fields (LBPS) \cite{knobelreiter2020belief}, a fully convolutional-densely connected neural network (FC-DCNN) \cite{hirner2021fc}, a
deep-learning assisted method to produce initial estimate with Semi-Global Block Matching method (SGBMP) \cite{hu2020deep}, a deep learning-based self-guided cost aggregation method (DSGCA) \cite{park2018deep}, a stereo matching algorithm with a pretrained network and global energy minimization SIGMRF \cite{nahar2017learned}, a multi-dimensional convolutional neural network (MSMD-ROB) \cite{lu2018cascaded} and a CNN-based network using ResNeXt (CBMBNet) \cite{chen2018crop}.
\begin{table}[!ht]
\renewcommand{\tabcolsep}{3pt}
\centering
\caption{Performance (Avg. err) of the FGS algorithm vs state-of-the-art \textit{learning-based disparity estimation algorithms} for stereo pairs in the Middlebury evaluation dataset version 3.0.  The weight of a stereo pair represents the level of difficulty in estimating disparity map from the respective stereo pair.  The algorithm with the highest performance is highlighted in bold.}
\label{tab:6}   
\begin{tabular}{ccccccccc}
\hline\noalign{\smallskip}
   \rot{90}{Images (Weight)} & \rot{90}{FGS} & \rot{90}{FC-DCNN \cite{hirner2021fc}} & \rot{90}{LBPS \cite{knobelreiter2020belief}} & \rot{90}{SGBMP \cite{hu2020deep}} &  \rot{90}{DSGCA \cite{park2018deep}} & \rot{90}{SIGMRF \cite{nahar2017learned}} & \rot{90}{MSMD-ROB \cite{lu2018cascaded}} & \rot{90}{CBMBNet \cite{chen2018crop}} \\
    \noalign{\smallskip}\hline\noalign{\smallskip}

Adiron $(8)$	&	3.63	&	2.87	&	1.92	&	6.50	&	7.68	&	3.07	&	2.85	&	\textbf{1.63}	\\
ArtL $(8)$	&	\textbf{4.81}	&	6.30	&	7.02	&	9.33	&	21.70	&	7.83	&	8.58	&	8.89	\\
Jadepl $(8)$	&	26.72	&	32.70	&	\textbf{24.9}	&	56.80	&	45.00	&	32.80	&	45.10	&	27.70	\\
Motor $(8)$	&	\textbf{3.36}	&	4.65	&	4.12	&	4.04	&	10.60	&	5.83	&	5.12	&	4.19	\\
MotorE $(8)$	&	5.90	&	4.58	&	\textbf{4.09}	&	5.43	&	10.40	&	5.92	&	4.99	&	4.12	\\
Piano $(8)$	&	4.23	&	4.45	&	\textbf{3.02}	&	4.77	&	11.50	&	5.38	&	3.75	&	3.22	\\
PianoL $(4)$	&	7.10	&	9.25	&	\textbf{3.63}	&	14.80	&	24.50	&	8.13	&	7.18	&	5.40	\\
Pipes $(8)$	&	7.71	&	10.00	&	\textbf{7.37}	&	7.85	&	19.90	&	11.30	&	11.00	&	8.03	\\
Playrm $(4)$	&	5.30	&	6.15	&	\textbf{4.83}	&	7.62	&	24.60	&	5.66	&	6.86	&	5.96	\\
Playt $(4)$	&	\textbf{3.19}	&	9.60	&	3.20	&	10.60	&	34.50	&	13.40	&	9.74	&	5.69	\\
PlaytP $(8)$	&	\textbf{3.05}	&	3.26	&	3.39	&	3.78	&	14.80	&	4.26	&	9.32	&	3.89	\\
Recyc $(8)$	&	3.71	&	2.67	&	1.71	&	3.19	&	7.56	&	3.07	&	2.74	&	\textbf{1.7}	\\
Shelvs $(4)$	&	5.30	&	10.00	&	\textbf{3.19}	&	5.00	&	17.30	&	8.57	&	3.56	&	7.70	\\
Teddy $(8)$	&	\textbf{1.9}	&	2.17	&	2.33	&	3.35	&	12.20	&	2.76	&	3.02	&	4.55	\\
Vintage $(4)$	&	10.22	&	9.34	&	\textbf{3.18}	&	30.00	&	43.80	&	15.50	&	9.59	&	5.71	\\
\hline
\textbf{Weighted}\\
\textbf{Average} &	6.45	&	7.67	&	\textbf{5.51}	&	10.38	&	18.70	&	8.63	&	9.19	&	6.65	\\
								
\noalign{\smallskip}\hline\noalign{\smallskip}
\end{tabular}
\end{table}

Avg. err metric for the FGS algorithm and the learning-based disparity estimation procedures are presented in Table~\ref{tab:6}.  When compared to the learning-based disparity estimation procedures, the FGS method provided the second-lowest weighted average of Avg. err of 6.45, provided the lowest Avg. err for 4 out of 10 difficult stereo pairs and the lowest Avg. error for 1 out of 5 moderate difficulty stereo pairs. 
%
\section{Conclusions}
\label{Conclusions}
We have presented a new probabilistic factor-graph-based disparity estimation algorithm that improves the accuracy of disparity estimates in stereo image pairs with varying texture and illumination characteristics by enforcing spatial dependencies among scene characteristics as well as among disparity estimates. In contrast to MRF models, our factor graph formulation allows a larger as well as a spatially variable neighborhood system dependent only on the local scene characteristics.  Our factor graph formulation can be used for obtaining \textit{maximum a posteriori} estimates from models or optimization problems with complex dependency structure among hidden variables.  The strategies of using \textit{a priori} distributions with shorter support and spatial dependencies are useful for improving the computational speed of message convergence in factor graph-based inference problems.  We rigorously evaluated the performance of the new factor-graph-based disparity estimation algorithm using Middlebury benchmark stereo datasets \cite{scharstein2003high}, \cite{scharstein2007learning}, \cite{hirschmuller2007evaluation}, and \cite{scharstein2014high}. Our experimental results indicate that the factor-graph algorithm provides disparity estimates with higher accuracy when compared to recent non-learning as well as learning-based disparity estimation algorithms using Middlebury evaluation dataset version 3.0 \cite{scharstein2002taxonomy}. The factor-graph algorithm may also be useful for other dense estimation problems such as optical flow estimation.



%
%
%
%
\bibliographystyle{spmpsci}     
\bibliography{ref.bib}

\end{document}